\pdfoutput=1

\documentclass[11pt]{article}

\usepackage{acl}
\usepackage{graphicx}
\usepackage{times}
\usepackage{latexsym}
\usepackage{placeins}
\usepackage{graphicx}
\usepackage{hyperref}
\usepackage{amsmath}
\usepackage{amssymb}

\usepackage[T1]{fontenc}

\usepackage[utf8]{inputenc}
\usepackage{comment}
\usepackage{microtype}

\title{MANTIS at \#SMM4H 2023: Leveraging Hybrid and Ensemble Models for Detection of Social Anxiety Disorder on Reddit}

\author{Sourabh Zanwar \\
  RWTH Aachen University \\
  \texttt{sourabh.zanwar@rwth-aachen.de} \And
  Daniel Wiechmann \\
  University of Amsterdam \\
  \texttt{d.wiechmann@uva.nl} \AND
  Yu Qiao \\ 
  RWTH Aachen University \\
  \texttt{yu.qiao@rwth-aachen.de} \\ \And
  Elma Kerz \\
  RWTH Aachen University \\
  \texttt{elma.kerz@ifaar.rwth-aachen.de} \\
  }

\begin{document}

\setlength{\abovedisplayskip}{7pt}
\setlength{\belowdisplayskip}{7pt}
\setlength{\abovedisplayshortskip}{0pt}
\setlength{\belowdisplayshortskip}{0pt}

\maketitle

\begin{abstract}

This paper presents our system employed for the Social Media Mining for Health 2023 Shared Task 4: Binary classification of English Reddit posts self-reporting a social anxiety disorder diagnosis. We systematically investigate and contrast the efficacy of hybrid and ensemble models that harness specialized medical domain-adapted transformers in conjunction with BiLSTM neural networks. The evaluation results outline that our best performing model obtained 89.31\% F1 on the validation set and 83.76\% F1 on the test set.

\end{abstract}

\FloatBarrier

\section{Introduction}
\vspace{-2mm}
According to the Anxiety \& Depression Association of America\footnote{\url{https://adaa.org/understanding-anxiety/facts-statistics}}, anxiety disorders rank as the most prevalent mental illnesses in the United States. An estimated 40 million adults, constituting 19.1\% of the population aged 18 and above, grapple with these conditions annually. This challenge is compounded by the scarcity of accessible mental health care services and the frequent occurrence of misdiagnoses, often causing individuals to unknowingly endure these disorders \cite{kasper2006anxiety}.


Natural Language Processing in combination with Machine Learning is increasingly recognized as having transformative potential to support healthcare professionals and stakeholders in the early detection, treatment and prevention of mental disorders \citep{zhang2022}. In this paper, we report on our participation in The Social Media Mining for Health Applications (\#SMM4H) 2023 workshop, which aims to promote automated methods for mining social media data for health informatics. We chose to participate in the Shared Task 4 competition, which was to improve social anxiety detection in Reddit posts \citep{klein2023}. We approached this task by developing hybrid and ensemble models combining domain-matched transformers with Bidirectional Long Short-Term Memory (BiLSTM) networks trained on a comprehensive set of engineered linguistic features. This set encompasses measures of morpho-syntactic complexity, lexical sophistication/diversity, readability, stylistics measures (register-specific ngram frequencies) and sentiment/emotion lexicons.
\vspace{-2mm}
\section{Data}
\vspace{-2mm}
The data for Task 4 consisted of 8117 Reddit posts written by users aged between 12 and 25 years.
These data were split into training (75\%), validation (8.4\%), and testing sets (16.6\%).
In preparation for model training, all texts were subjected to preprocessing procedures including eliminating HTML, URLs, excessive spaces, and emojis from the text, as well as rectifying inconsistent punctuation.

\vspace{-2mm}
\section{System Description}
\vspace{-2mm}
Our systems leveraged three domain-adapted Transformer-based pretrained language models (PLM), a BiLSTM trained on engineered features and their combination forming into hybrid and ensemble models. Domain-adapted pretrained language models include: (1) PsychBERT \cite{vajre2021psychbert} (2) Mental RoBERTa \cite{ji-etal-2022-mentalbert} and (3) Clinical BERT \cite{alsentzer-etal-2019-publicly}. All PLMs were obtained from the Huggingface \cite{wolf2020Transformers}, choosing the uncased, where applicable, base versions. We constructed a BiLSTM trained on 168 features that fall into six categories. All measurements of these features were obtained using a system that employs a sliding window technique to compute sentence-level measurements. The BiLSTM model is formulated as:\par
\vspace{-3mm}
{\small
\setlength{\abovedisplayskip}{3pt}
\setlength{\belowdisplayskip}{\abovedisplayskip}
\begin{align*}
    h_N^{(L)} & = \text{BiLSTM}^{L}_{H}(\text{CM}_1^N), h^{(L)}_N = h_{f,N}^{(L)}\oplus h_{b,N}^{(L)}\\[-0.3em]
    \hat{y} &= \text{Softmax}\bigl(Wfc_2\bigl(fc_1(h_N^{(L)})\bigr) + b\bigr)
\end{align*}
}%
where $fc_i(x)=\text{ReLU}(W_ix + b_i)$, $\text{BiLSTM}_{H}^{L}(\cdot)$ is a $L$ layer BiLSTM with hidden size of $H$. $\text{CM}_1^N=(\text{CM}_1, \text{CM}_2\dots, \text{CM}_N)$, where $\text{CM}_i$ represents the linguistic features for the $i$th sentence of a post consisting of $N$ sentences. 
The last hidden representation of the last layer in forward and backward directions are denoted by ${h_{f,N}^{(L)}}$ and ${h_{b,N}^{(L)}}$. $\oplus$ denotes the concatenation operator. 

The hybrid model combines a Mental RoBERTa model with above BiLSTM.\par
\vspace{-3mm}
{\small
\setlength{\abovedisplayskip}{3pt}
\setlength{\belowdisplayskip}{\abovedisplayskip}
\begin{align*}
    S_1^M & = \text{Mental\_RoBERTa}(T_1^M) \\[-0.3em]
    s_M^{(L_1)} & = \text{BiLSTM}^{L_1}_{H_1}(S_1^M), s^{(L_1)}_M = s_{f,M}^{(L_1)}\oplus s_{b,M}^{(L_1)} \\[-0.3em]
    h_N^{(L_2)} & = \text{BiLSTM}^{L_2}_{H_2}(\text{CM}_1^N), h^{(L_2)}_N = h_{f,N}^{(L_2)}\oplus h_{b,N}^{(L_2)} \\[-0.3em]
    \hat{y} &= \ \text{Softmax}(Wfc_3\bigl[fc_1(s_M^{(L_1)})\oplus fc_2(h_N^{(L_2)})\bigr]+b)
\end{align*}
}%
where $T_1^M$ is the sequence of tokens from a post. 

We constructed three distinct ensemble models using the stacking method: ensemble model (1) composed of instances from the hybrid model, which emerged as the most accurate base model (M6), (2) combining hybrid models with fine-tuned PsychBERT models (M7), and (3) consisting of Mental RoBERTa models, PsychBERT models, and BiLSTM models (M8). The resulting models represent homogeneous ensemble (HOE), intermediate and heterogeneous ensemble (HEE) approaches \cite{ganaie2022ensemble}. 
As meta-learners, Support Vector Classifer, Logistic Regression, Gradient Boosting and Ridge Regression and XGBoost were used.

Further details are provided in the supplementary material (\href{https://drive.google.com/drive/folders/1s5veJWWyuFBiQu9Vx3Z5LM0uIosO199f?usp=sharing}{https://shorturl.at/epuF3}).

\vspace{-2mm}
\section{Results and Evaluation}
\vspace{-2mm}
The results of our models on the validation set and test set are presented in Table \ref{tab:valresults}. The Mental RoBERTa model achieved the highest performance (F1=86.59\%) among the PLMs, outperforming the PsychBERT and ClinicalBERT models by 4\% and 14.73\%, respectively. This finding indicates that the detection of anxiety on Reddit sees a marked improvement from pretraining the PLM on mental health-related subreddits, as opposed to pretraining on clinical text. The hybrid models consistently outperformed the standalone PLM across all model iterations, yielding an average increase in F1 scores of 0.3\%. The use of model stacking enhanced classification outcomes with performance boosts ranging between 1.86\% and 2.44\% in F1 score. The highest balanced classification score was achieved by the HEE model (M8). A variant of this model using ridge regression as a meta-learner (M12) achieved the best performance on the test set (F1 = 83.76\%, mean\textsubscript{all\ teams} = 79.3\%, median\textsubscript{all\ teams} = 82.4\%). The HOE model (M6) achieved the second-highest performance and the best precision among all models examined. This suggests that both ensemble approaches can produce beneficial, albeit distinct, impacts on the detection of social anxiety disorder.

%


\begin{table}[]
    \caption{Results on the validation set (top) and test set (bottom). For each ensemble model, we report results of the best performing meta-learner.}
    \label{tab:valresults}
    \setlength{\tabcolsep}{2pt}
    \centering
    \small
    \vspace{-2mm}
\begin{tabular}{|l|l|l|l|}
\hline
\textbf{Detection   model}                                                                       & \textbf{F1} & \textbf{P} & \textbf{R} \\
\hline
\textit{Pretrained Language Models}                                                        &             &                    &                 \\
M1: Mental RoBERTa & 86.59 & 81.85 & 91.91 \\
M2: PsychBERT & 82.59 & 79.46 & 85.98 \\
M3: ClinicalBERT & 71.86 & 69.36 & 74.55 \\ \hline
M4: BiLSTM & 59.01 & 60.03 & 58.92 \\ \hline
M5:  Hybrid Model & 86.87 & 83.30 & 90.76 \\ \hline
\textit{Ensemble Models}                                                                         &             &                    &                 \\
M6: HOE M5 (GB) & 88.80 & 85.57 & 92.28 \\
M7: HEE M5+M2  (GB) & 88.73 & 84.90 & 92.92 \\
M8: HEE M1+M2+M4 (GB) & 89.31 & 85.02 & 94.06 \\
\hline
\hline
Test set                                                                      & \textbf{F1} & \textbf{P} & \textbf{R} \\
\hline

M6: HOE: M5 (GB)                       &   83.63          &  \textbf{81.07}                  &    86.35             \\
   \hline
M12: HEE M1+M2+M4 (Ridge) &    \textbf{83.76}

         &    80.6                &       \textbf{87.17}          \\
         \hline
\end{tabular}
\vspace{-5mm}
\end{table}

\vspace{-2mm}



\FloatBarrier
\bibliography{custom, supplementary}

\begin{thebibliography}{8}
\expandafter\ifx\csname natexlab\endcsname\relax\def\natexlab#1{#1}\fi

\bibitem[{Alsentzer et~al.(2019)Alsentzer, Murphy, Boag, Weng, Jindi, Naumann,
  and McDermott}]{alsentzer-etal-2019-publicly}
Emily Alsentzer, John Murphy, William Boag, Wei-Hung Weng, Di~Jindi, Tristan
  Naumann, and Matthew McDermott. 2019.
\newblock Publicly available clinical {BERT} embeddings.
\newblock In \emph{Proceedings of the 2nd Clinical Natural Language Processing
  Workshop}, pages 72--78. Association for Computational Linguistics.

\bibitem[{Ganaie et~al.(2022)Ganaie, Hu, Malik, Tanveer, and
  Suganthan}]{ganaie2022ensemble}
Mudasir~A Ganaie, Minghui Hu, AK~Malik, M~Tanveer, and PN~Suganthan. 2022.
\newblock Ensemble deep learning: A review.
\newblock \emph{Engineering Applications of Artificial Intelligence},
  115:105151.

\bibitem[{Ji et~al.(2022)Ji, Zhang, Ansari, Fu, Tiwari, and
  Cambria}]{ji-etal-2022-mentalbert}
Shaoxiong Ji, Tianlin Zhang, Luna Ansari, Jie Fu, Prayag Tiwari, and Erik
  Cambria. 2022.
\newblock {M}ental{BERT}: Publicly available pretrained language models for
  mental healthcare.
\newblock In \emph{Proceedings of the Thirteenth Language Resources and
  Evaluation Conference}, pages 7184--7190, Marseille, France. European
  Language Resources Association.

\bibitem[{Kasper(2006)}]{kasper2006anxiety}
Siegfried Kasper. 2006.
\newblock Anxiety disorders: under-diagnosed and insufficiently treated.
\newblock \emph{International Journal of Psychiatry in Clinical Practice},
  10(sup1):3--9.

\bibitem[{Klein et~al.(2023)Klein, Banda, Guo, Flores~Amaro, Rodriguez-Esteban,
  Sarker, Schmidt, Xu, and Gonzalez-Hernandez}]{klein2023}
AZ~Klein, JM~Banda, Y~Guo, JI~Flores~Amaro, R~Rodriguez-Esteban, A~Sarker,
  AL~Schmidt, D~Xu, and G~Gonzalez-Hernandez. 2023.
\newblock Overview of the eighth social media mining for health applications
  (\#smm4h) shared tasks at the {AMIA} 2023 annual symposium.
\newblock \emph{Proceedings of the Eighth Social Media Mining for Health
  Applications (\#SMM4H) Workshop and Shared Task}.

\bibitem[{Vajre et~al.(2021)Vajre, Naylor, Kamath, and
  Shehu}]{vajre2021psychbert}
Vedant Vajre, Mitch Naylor, Uday Kamath, and Amarda Shehu. 2021.
\newblock Psychbert: a mental health language model for social media mental
  health behavioral analysis.
\newblock In \emph{2021 IEEE International Conference on Bioinformatics and
  Biomedicine (BIBM)}, pages 1077--1082. IEEE.

\bibitem[{Wolf et~al.(2020)Wolf, Chaumond, Debut, Sanh, Delangue, Moi, Cistac,
  Funtowicz, Davison, Shleifer et~al.}]{wolf2020Transformers}
Thomas Wolf, Julien Chaumond, Lysandre Debut, Victor Sanh, Clement Delangue,
  Anthony Moi, Pierric Cistac, Morgan Funtowicz, Joe Davison, Sam Shleifer,
  et~al. 2020.
\newblock Transformers: State-of-the-art natural language processing.
\newblock In \emph{Proceedings of the 2020 Conference on Empirical Methods in
  Natural Language Processing: System Demonstrations}, pages 38--45.

\bibitem[{Zhang et~al.(2022)Zhang, Schoene, and Ananiadou}]{zhang2022}
T.~Zhang, A~Schoene, and S.~Ananiadou. 2022.
\newblock Natural language processing applied to mental illness detection: A
  narrative review.
\newblock \emph{NPJ Digital Medicine}, 5:46.

\end{thebibliography}
\bibliographystyle{acl_natbib}

\appendix

\FloatBarrier
\newpage

\end{document}